\documentclass[conference]{IEEEtran}
\IEEEoverridecommandlockouts
\usepackage{cite}
\usepackage{amsmath,amssymb,amsfonts}
\usepackage{graphicx}
\usepackage{textcomp}
\usepackage{xcolor}
\usepackage{booktabs}
\usepackage{url}
\usepackage{multirow}
\usepackage{tablefootnote}
\usepackage{ulem}
\usepackage{algorithm}
\usepackage{algpseudocode}

\def\BibTeX{{\rm B\kern-.05em{\sc i\kern-.025em b}\kern-.08em
    T\kern-.1667em\lower.7ex\hbox{E}\kern-.125emX}}

\begin{document}

\title{Tracing Footsteps of Similar Cities: Modeling Urban Economic Vitality with Dynamic Inter-City Graph Embeddings}

% 方法一：修改原有代码，让后两个作者居中
% \author{
% \begin{tabular}{@{}c@{\hspace{1.2cm}}c@{\hspace{1.2cm}}c@{}}
% \begin{minipage}[t]{5.2cm}
% \centering
% \textbf{Xiaofeng Li}\textsuperscript{*} \\
% %School of Information Science and Technology \\
% ShanghaiTech University \\
% Shanghai, China \\
% lixf2022@shanghaitech.edu.cn
% \end{minipage} &
% \begin{minipage}[t]{5.2cm}
% \centering
% \textbf{Xiangyi Xiao}\textsuperscript{*} \\
% %School of Information Science and Technology \\
% ShanghaiTech University \\
% Shanghai, China \\
% xiaoxy2025@shanghaitech.edu.cn
% \end{minipage} &
% \begin{minipage}[t]{5.2cm}
% \centering
% \textbf{Xiaocong Du} \\
% %School of Information Science and Technology \\
% ShanghaiTech University \\
% Shanghai, China \\
% duxc2023@shanghaitech.edu.cn
% \end{minipage} \\[5em]

% % 后两个作者居中，使用multicolumn跨越三列
% \multicolumn{3}{c}{
% \begin{tabular}{@{}c@{\hspace{2.5cm}}c@{}}
% \begin{minipage}[t]{5.2cm}
% \centering
% \textbf{Ying Zhang} \\
% %School of Information Science and Technology \\
% ShanghaiTech University \\
% Shanghai, China \\
% sgjzp.joyce@gmail.com
% \end{minipage} &
% \begin{minipage}[t]{5.2cm}
% \centering
% \textbf{Haipeng Zhang}\textsuperscript{†} \\
% %School of Information Science and Technology \\
% ShanghaiTech University \\
% Shanghai, China \\
% zhanghp@shanghaitech.edu.cn
% \end{minipage}
% \end{tabular}
% } \\
% \end{tabular}
% }
\author{
\IEEEauthorblockN{
Xiaofeng Li\IEEEauthorrefmark{1}$^{*}$,
Xiangyi Xiao\IEEEauthorrefmark{1}$^{*}$,
Xiaocong Du\IEEEauthorrefmark{1},
Ying Zhang\IEEEauthorrefmark{1},
and Haipeng Zhang\IEEEauthorrefmark{1}$^{\dagger}$
}
\IEEEauthorblockA{\IEEEauthorrefmark{1}ShanghaiTech University, Shanghai, China\\
%Email: lixf2022@shanghaitech.edu.cn, xiaoxy2025@shanghaitech.edu.cn, duxc2023@shanghaitech.edu.cn, sgjzp.joyce@gmail.com, zhanghp@shanghaitech.edu.cn}
Email: lixf2022, xiaoxy2025, duxc2023@shanghaitech.edu.cn, \\
sgjzp.joyce@gmail.com, zhanghp@shanghaitech.edu.cn
}
\thanks{$^{*}$ These authors contributed equally to this work.}
\thanks{$^{\dagger}$ Corresponding author.}
}

\maketitle
\begin{abstract}
Urban economic vitality is a crucial indicator of a city’s long-term growth potential, comprising key metrics such as the annual number of new companies and the population employed. However, modeling urban economic vitality remains challenging. This study develops ECO-GROW, a multi-graph framework modeling China's inter-city networks (2005-2021) to generate urban embeddings that model urban economic vitality. Traditional approaches relying on static city-level aggregates fail to capture a fundamental dynamic: the developmental trajectory of one city today may mirror that of its structurally similar counterparts tomorrow. ECO-GROW overcomes this limitation by integrating industrial linkages, POI similarities, migration similarities and temporal network evolution over 15 years. The framework combines a Dynamic Top-K GCN to adaptively select influential inter-city connections and an adaptive Graph Scorer mechanism to dynamically weight cross-regional impacts. Additionally, the model incorporates a link prediction task based on Barabasi Proximity, optimizing the graph representation. Experimental results demonstrate ECO-GROW's superior accuracy in predicting \textit{entrepreneurial activities and employment trends} compared to conventional models. By open-sourcing our code, we enable government agencies and public sector organizations to leverage big data analytics for evidence-based urban planning, economic policy formulation, and resource allocation decisions that benefit society at large.

\end{abstract}


\section{Introduction}
%%%%%%%%%%%%%%%%%%%%%%%%%%%%%%%%%%%%%%%%%%%%%%%%%%%%%%%%%%%%%%%%%%%%%%%%%%%%%
% 1 本质：the developmental trajectory of one city today may mirror that of its structurally similar counterparts tomorrow 这段内容和标题tracing footsteps关联度高，所以放在第一个，顺便引出图1.
%   baseline: HREP / MVURE / MGFN 依赖于静态的城市内数据。
%   ECO-GROW: capture Inter-city relationships and temporal evolution 
%----------------------------------------------------------------------------%
% 2 本质：影响经济的一个非常重要因素：产业结构，捕捉城市间工业相似性。
%   必要的铺垫：引入工业数据，所以还是得把引入六种网络提前说了，但是简单说。
%   baseline：node2vec, GAE, HREP, MVURE, MGFN 通常侧重于捕捉和重建不同的城市属性或图本身的结构信息，没有捕捉动态产业组合的演化。
%   ECO-GROW：经济的增长是通过升级生产和出口的产品类型来实现的(原barabasi文章中的说法)。 模型将这一概念从产品之间的邻近度扩展到了城市之间的工业组合相似性。基于 Barabási Proximity 的链接预测任务作为其监督学习目标之一 利用复杂的领域特定先验知识（复杂网络理论中关于产业转移模式和经济依赖性）来指导嵌入学习economic vitality。
%----------------------------------------------------------------------------%
% 3 本质：muti-graph中不同图有自己的attributes，捕捉这种图的特异性才能更好利用邻居信息。
%   baseline：HREP, MVURE, MGFN 对每张图使用固定的k,无法根据图的特点做出适应性调整。
%   ECO-GROW: top-k gcn模块中为每种类型的图auto-select k个邻居,增强了模型在处理不同图数据时的灵活性.
%%%%%%%%%%%%%%%%%%%%%%%%%%%%%%%%%%%%%%%%%%%%%%%%%%%%%%%%%%%%%%%%%%%%%%%%%%%%%%%

\begin{figure*}[htbp!]
\vspace{-6mm}
\centering
\includegraphics[width=0.9\textwidth]{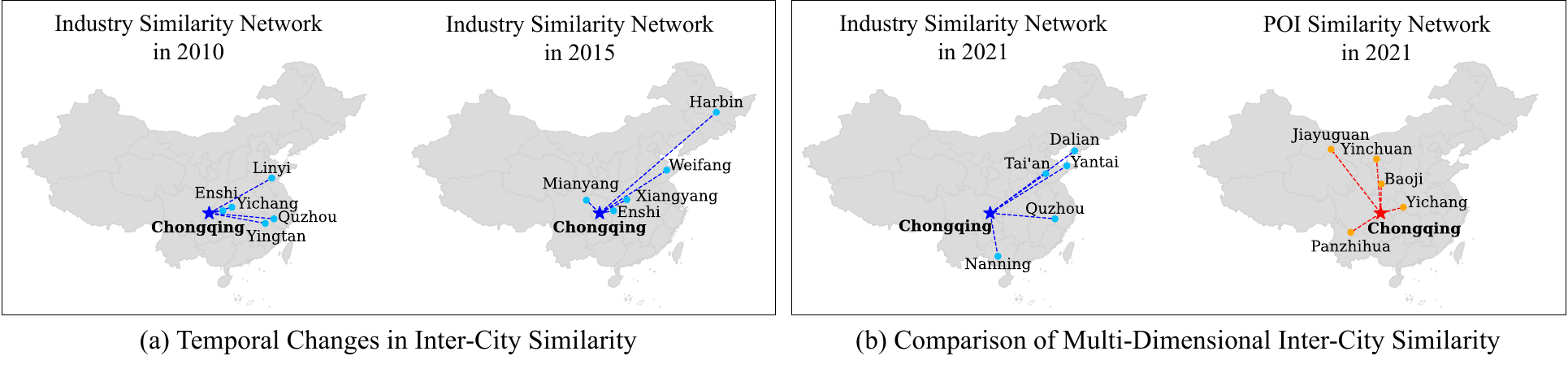}
\vspace{-3mm}
\caption{The maps illustrate industry similarity dynamics and differences in city similarity networks, focusing on a part of China. The star marks Chongqing and the circles indicate the top five similar cities. (a) Shows changes in industry similarity over time. (b) Compares the top five similar cities across attributes.}
\label{fig:graph}
\vspace{-6mm}
\end{figure*}

Urban economic vitality—reflecting a city’s overall economic health and growth potential~\cite{10825750}—is typically measured by two core indicators: the annual number of new companies and the population employed~\cite{Li2020Quantitative}. These factors, though distinct, are interconnected and serve as key measures of a city’s economic dynamics and growth potential~\cite{Lv2020Evaluation}. Predicting these indicators remains challenging due to the complex, dynamic nature of urban economies. Existing approaches often rely on static, city-level aggregates, neglecting the evolving relationships between cities and their developmental trajectories~\cite{zhang2021multi,wu2022multi, zhou2023heterogeneous}. However, cities are not isolated entities. The economic growth of one city can be influenced by and serve as a reference for structurally or geographically similar cities. For instance, the Silicon Valley effect exemplifies how knowledge spillovers, fueled by research investment, skilled labor, and academic collaboration, lead to concentrated innovation in a specific region~\cite{audretsch1996r}.
%For instance, in China, the coastal regions rely on resources and products from inland areas, forming a unique interdependent regional development model that drives national economic expansion~\cite{zhang2020uneven}.
We also illustrate the changing inter-city relationships using Chongqing City as an example (Figure~\ref{fig:graph}(a)). Cities with high industry similarity shifted from coastal regions in 2010 to larger northeastern cities in later years, reflecting the dynamic process of industrial convergence noted in prior studies on China's economic development~\cite{li2014spatial_correlation}. This observation highlights that modeling inter-city relationships together with their temporal evolution is essential for understanding and predicting urban economic vitality.

%%%%%%%%%%%%%%%%%%% Figure: map %%%%%%%%%%%%%%%%%%%
%%%%%%%%%%%%%%%%%%% Figure: map %%%%%%%%%%%%%%%%%%%

% 1 本质：the developmental trajectory of one city today may mirror that of its structurally similar counterparts tomorrow 这段内容和标题tracing footsteps关联度高，所以放在第一个，顺便引出图1.
%   baseline: HREP / MVURE / MGFN 依赖于静态的城市内数据。
%   ECO-GROW: capture Inter-city relationships and temporal evolution 

Traditional models often embed cities based only on local structures or attributes~\cite{kipf2016variational, zhang2021multi, wu2022multi}, overlooking broader inter-city economic linkages that drive parallel growth among cities. To address this limitation, we introduce a link prediction objective guided by Barabasi Proximity—a concept adapted from Barabasi’s product space theory~\cite{barabasi2007productspace}—to measure the likelihood of economic connections between cities based on industrial similarity. By leveraging domain-specific prior knowledge on industrial transfer patterns and economic dependencies, our framework learns embeddings that are more sensitive to inter-city economic relationships and better capture the structural dynamics of urban economies.

% 2 本质：影响经济的一个非常重要因素：产业结构，捕捉城市间工业相似性。
%   必要的铺垫：引入工业数据，所以还是得把引入六种网络提前说了，但是简单说。
%   baseline：node2vec, GAE, HREP, MVURE, MGFN 通常侧重于捕捉和重建不同的城市属性或图本身的结构信息，没有捕捉动态产业组合的演化。
%   ECO-GROW：经济的增长是通过升级生产和出口的产品类型来实现的(原barabasi文章中的说法)。 模型将这一概念从产品之间的邻近度扩展到了城市之间的工业组合相似性。基于 Barabási Proximity 的链接预测任务作为其监督学习目标之一 利用复杂的领域特定先验知识（复杂网络理论中关于产业转移模式和经济依赖性）来指导嵌入学习economic vitality。

To comprehensively model urban economic vitality,
it is essential to incorporate critical urban attributes.
%While early studies are limited to using statistical indicators
Beyond traditional statistical indicators ~\cite{audretsch2004entrepreneurship,acs2006entrepreneurship}, 
%recent advancements in urban embedding learning have begun to use more complex attributes like population mobility and infrastructure
and recent mobility networks~\cite{zhang2021multi, wu2022multi, zhou2023heterogeneous},
%by incorporating corresponding networks.
%In addition to these networks, 
our framework introduces another two critical networks, regional economic connections and industrial structure~\cite{Cheng2019Source,Xia2021The} (mentioned before in the Chongqing City case), which serves as a determinant and a consequence of economic growth~\cite{Zhu2021industrial}.

% 3 本质：muti-graph中不同图有自己的attributes，捕捉这种图的特异性才能更好利用邻居信息。
%   baseline：HREP, MVURE, MGFN 对每张图使用固定的k,无法根据图的特点做出适应性调整。
%   ECO-GROW: top-k gcn模块中为每种类型的图auto-select k个邻居,增强了模型在处理不同图数据时的灵活性
We then focus on distinctive urban association representations under various attribute networks, as well as attribute fusion.
As illustrated in Figure~\ref{fig:graph} (b), cities with high industry similarity can exhibit different POI similarity patterns, suggesting that
different graphs capture heterogeneous structures.
%, and capturing these differences is essential for effectively utilizing neighboring city information.
Therefore, to effectively utilize neighboring city information, our framework employs a Dynamic Top-K GCN (DTKGCN) module that dynamically adjusts the number of neighbors, k, for each graph type.
Unlike models that apply a fixed k value for all graphs~\cite{zhou2023heterogeneous, zhang2021multi}, the DTKGCN module allows the model to select the most relevant neighboring cities based on the unique characteristics of each graph (e.g., industry, POI, mobility networks). To further enhance the model's ability to handle diverse networks, we introduce the Graph Scorer mechanism, which dynamically assigns weights to each static graph, ensuring that the most relevant graphs have a stronger influence on the final embedding.

Building on these insights, we propose \textbf{ECO-GROW} (\textbf{E}mbedding of \textbf{C}ities through \textbf{O}ptimized \textbf{G}raph \textbf{R}epresentation \textbf{O}f \textbf{W}eighted neighbors), a framework for modeling urban economic vitality through multi-graph learning.
It integrates six inter-city networks to capture diverse urban attributes, employs a Dynamic Top-K GCN (DTKGCN) to adaptively select the most relevant neighbors for each network, and uses a Graph Scorer to balance their contributions.
Guided by domain-specific prior knowledge from Barabasi’s complex network theory, ECO-GROW learns optimized graph representations that enable accurate prediction of new companies and employment trends, demonstrating strong practical value.

In summary, our contributions are summarized as follows:
\begin{itemize}
    \item We propose a novel multi-graph framework ECO-GROW to model economic vitality representations, integrating dynamic inter-city information and static multi-network structures. By incorporating key modules like DTKGCN and Graph Scorer, our framework demonstrates superior performance in downstream tasks.
    
    \item We use domain-specific prior knowledge to supervise model training, enabling the model to better capture intricate economic relationships among cities. 
    As a result, ECO-GROW outperforms all baselines, with significant improvements across all metrics, including a 20.46\% reduction in RMSE and a 5.96\% increase in R\(^2\).

    \item This work demonstrates the robustness of ECO-GROW by testing on various downstream tasks and different time periods. The model shows the potential for other economic tasks, such as predicting housing market trends or income distribution, offering insights for policy making.
    %\textcolor{blue}{Case study demonstrates our model's capability to predict stable economic conditions, which may also help identify anomalous economic phenomena that deviate from expected structural trends.} 
   We open-source the code\footnote{\url{https://github.com/sherry18510/ECO_GROW}} to foster research and  applications in urban planning and policymaking.
\end{itemize}

%In the following sections, we introduce the proposed model in detail, discuss the experimental setup and results, review related work, and conclude with final remarks.

\section{Related Work}
\subsection{Graph Embedding}
Graph embedding techniques represent graph structures in low-dimensional spaces, facilitating tasks like node classification and link prediction. Node2Vec~\cite{grover2016node2vec} learns embeddings via biased random walks, preserving both local and global graph structures. Building on this, Graph Neural Networks (GNNs) leverage graph topologies and node features. The Graph Convolutional Network (GCN)~\cite{kipf2016semi} aggregates information from local neighborhoods, while the Graph Attention Network (GAT)~\cite{velickovic2018graph} uses attention mechanisms to weigh neighbors' importance. Graph Autoencoder (GAE)~\cite{kipf2016variational} takes an unsupervised approach by reconstructing graph topology to capture latent representations. The advances in graph embedding offer powerful and insightful tools to enhance traditional region embedding approaches.

\subsection{Region Representation Learning}
Region representation learning focuses on modeling relationships between urban attributes, which are traditionally applied to tasks such as crime rate prediction, check-in frequency analysis, and land-use classification. Recent advancements have shifted toward integrating diverse properties through more sophisticated methods~\cite{yao2018representing,pixels2023progress,musecl2023,cgap2023}. For example, MVURE~\cite{zhang2021multi} employs a multi-view learning framework with adaptive weight assignment to fuse human mobility and region attributes, demonstrating strong performance in tasks like crime prediction. Similarly, MGFN~\cite{wu2022multi} utilizes a multi-graph fusion module and cross-attention mechanisms to capture relationships within mobility graphs. HREP~\cite{zhou2023heterogeneous} constructs a heterogeneous region graph by integrating human mobility, geographic proximity, and POI distributions while leveraging self-attention and prompt learning to align embeddings with downstream tasks. Inspired by these approaches, our work extends region representation learning to model spatiotemporal dynamics for urban economic forecasting, addressing the unique challenges of predicting entrepreneurial growth.

\section{Preliminaries}

\subsection{Multi-Graph Construction}
Cities are complex systems characterized by geographic, mobility, and industrial attributes. To model these relationships, we construct six networks $\mathcal{G} = \{ G_\text{dist}, G_\text{s}, G_\text{d}, G_\text{poi}, G_\text{ind}, G_\text{emp} \}$, each capturing a specific attribute. 
%For all these networks, the edge weight \( A_{ij} \) represents the similarity or relationship between two cities \( v_i \) and \( v_j \). The specific formula for each network's edge weight is described below.

\subsubsection{Distance Network $G_\text{dist}$}
constructed using the inverse of geographic distance between cities to reflect spatial closeness.

\subsubsection{Population Flow Network $G_\text{s}, G_\text{d}$}
built from migration data, where similarity is measured by the cosine similarity of source and destination distributions~\cite{zhang2021multi}.

\subsubsection{POI Similarity Network $G_\text{poi}$}
derived from the TF-IDF representations of POI categories in each city to capture similarity.

\subsubsection{Industry Similarity Network $G_\text{ind}$}
constructed in a similar way, comparing cities’ industrial compositions using TF-IDF and cosine similarity.

\subsubsection{Original Feature Clustering Network $G_\text{emp}$}
generated by clustering cities based on explicit socioeconomic features (e.g., GDP, population) using K-means.

\subsection{Problem Statement}
Given the city characteristics \( F \) and the set of correlation networks \( \mathcal{G} \), the goal of modeling urban economic vitality is to learn a distributed and low dimensional embedding of each city, which is denoted as \( \mathcal{E}^t \),
\[
\mathcal{E}^t = \{e_1^t, e_2^t, \ldots, e_n^t\}, \quad e_i^t \in \mathbb{R}^d, \, \forall i \in [1, n]
\]
where \( e_i^t \) denotes the embedding of city \( i \) in year \( t \), and \( d \) is the dimension of embedding.

\section{Methodology}
\begin{figure*}[ht!]
\vspace{-6mm}
\centering
\includegraphics[width=0.85\textwidth]{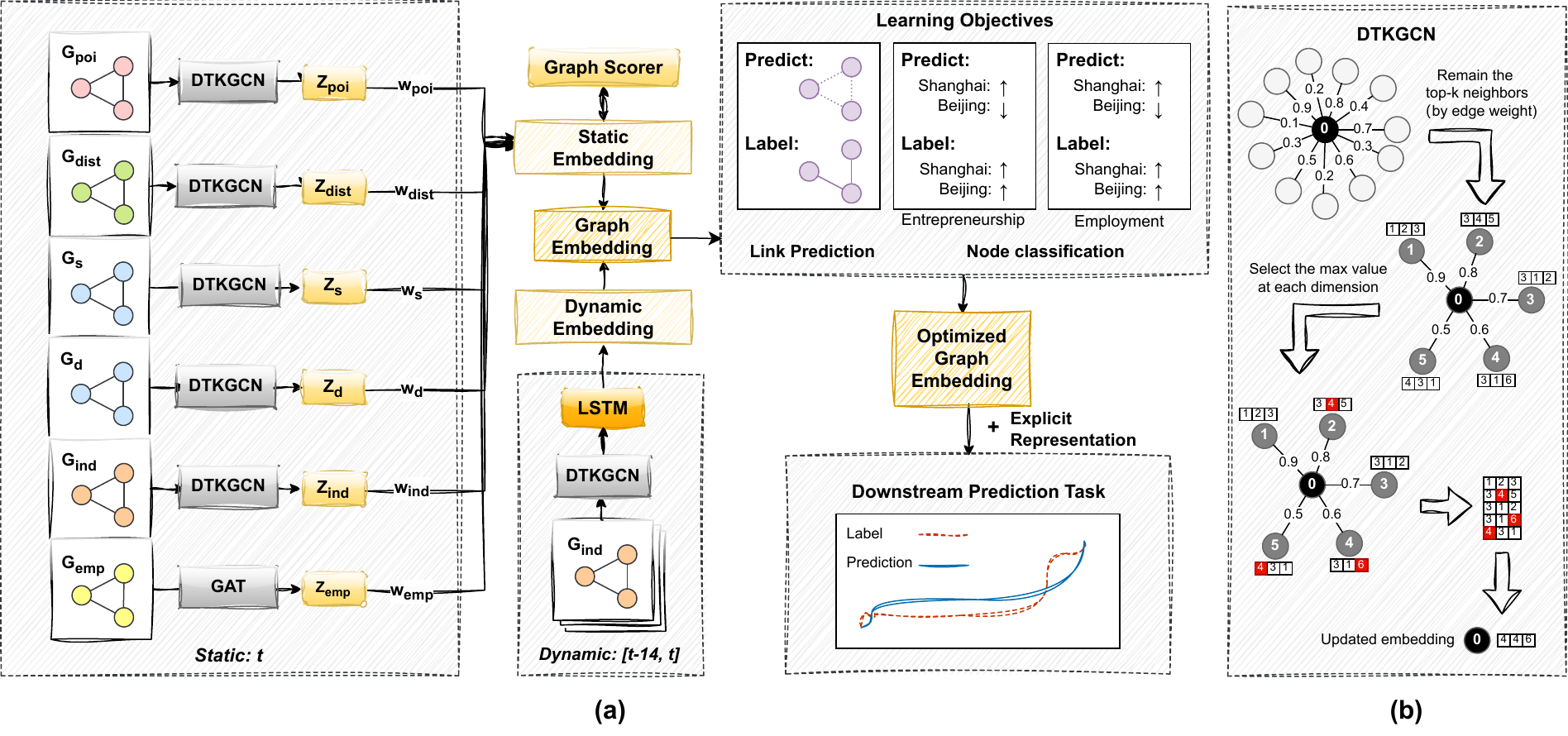}
\vspace{-3mm}
\caption{(a) The overall architecture of ECO-GROW. (b) DTKGCN Module: For each node in a certain graph $g$, the top-$k_g$ neighbors are selected, retaining the maximum value in each dimension to form the new embedding.}
\label{fig:model}
\vspace{-6mm}
\end{figure*}

%As shown in Figure~\ref{fig:model}(a), our model processes both static and dynamic inputs, constructed from six types of inter-city networks. For city graphs, we primarily use DTKGCN to aggregate information, where each graph learns its own dynamic $k$ value, with static graphs further processed using a Graph Scorer to adaptively learn graph embeddings. The model is trained with three learning objectives to capture economic vitality profiles.
This section introduces the proposed model ECO-GROW, with its 
overall framework shown in Figure~\ref{fig:model}(a). In the following sections, we will elaborate (1) DTKGCN, (2 Graph Scorer, (3) learning objectives, and (4) the overall process pipeline of the proposed ECO-GROW.

%%%%%%%%%%%%%%%%%%% Figure: model %%%%%%%%%%%%%%%%%%%
%%%%%%%%%%%%%%%%%%% Figure: model %%%%%%%%%%%%%%%%%%%

\subsection{DTKGCN}
In the field of Graph Neural Networks (GNNs), effective aggregation of information from neighboring nodes is fundamental to model performance. 
Traditional GCNs utilize uniform aggregation, which can dilute crucial information due to noise or sparsity~\cite{Peng2021Reinforced}.
Furthermore, many models rely on fixed neighborhood sizes, which may not capture the most relevant information, especially in dynamic graphs with heterogeneous structures.
To enhance this aggregation process, we introduce a mechanism called Dynamic Top-K GCN (DTKGCN), which adaptively selects the top-k most relevant neighbors based on the characteristics of each graph type, as shown in Figure~\ref{fig:model}(b).

%%%%%%%%%%%%%%%%%%% Figure: model %%%%%%%%%%%%%%%%%%%
% \begin{figure}[htbp]
% \centering
% \includegraphics[width=\linewidth]{figure/model/model_tkgcn.pdf}
% \caption{DTKGCN Module: For each node in a certain graph $g$, the top-$k_g$ neighbors are selected, retaining the maximum value in each dimension to form the new embedding.}
% \label{fig:tkgcn}
% \end{figure}
%%%%%%%%%%%%%%%%%%% Figure: model %%%%%%%%%%%%%%%%%%%
%\textcolor{red}{Unlike conventional methods where \(k\) is fixed, in our framework, each network learns a distinct \(k\) value tailored to the specific characteristics of the respective graph, reflecting the heterogeneous nature of urban data. For each graph \(g \in \{G_{\text{dist}}, G_{\text{poi}}, G_{s}, G_{d}, G_{\text{ind}}\}\), we introduce a dynamic $k_g$ corresponding to the graph \(g\). This allows the model to adjust $k_g$ based on the characteristics of the specific graph, enabling the model to adaptively select the most relevant neighboring nodes.}
The DTKGCN layer takes a node feature matrix \(X \in \mathbb{R}^{N \times F_\text{in}}\) and an adjacency matrix \(A \in \mathbb{R}^{N \times N}\), where \(N\) represents the number of nodes and \(F_\text{in}\) denotes the input feature dimension. The output is a new representation \(H_\text{agg} \in \mathbb{R}^{N \times F_\text{out}}\), where \(F_\text{out}\) is the output feature dimension.

Initially, the node features undergo a linear transformation:
\begin{equation}
	H = XW \in \mathbb{R}^{N \times F_\text{out}}
\end{equation}
where \(W \in \mathbb{R}^{F_\text{in} \times F_\text{out}}\) is the weight matrix, and \(H\) represents the transformed node features.

After this transformation, we move on to the neighbor aggregation process. For each graph \(g \in \{G_{\text{dist}}, G_{\text{poi}}, G_{s}, G_{d}, G_{\text{ind}}\}\), we introduce \(k_g\), a learnable parameter that determines the number of top neighbors to consider during the aggregation process. This parameter is optimized during the training process via gradient descent, where \(k_g\) is treated as a continuous parameter during backpropagation but discretized during forward aggregation.
Next, we compute edge weights from the adjacency matrix to identify the top \(k_g\) neighbors for each node in graph $g$. For a given node \(i\) in graph \(g\), the following operation is performed:
\begin{equation}
	\mathcal{I}_{\text{top-$k_g$}} = \text{top-$k_g$}(A_g[i], k_g)
\end{equation}
The \(\text{top-$k_g$}\) function returns indices of the top \(k_g\) neighbors based on their edge weights in $A_g[i]$.

The aggregation of features is conducted solely from the selected top \(k_g\) neighbors. The aggregation process for each node \(i\) is defined as follows:
\begin{equation}
	H_{\text{agg}}[i] = \max_{j \in \mathcal{I}_{\text{top-$k_g$}}[i]} H[j]
\end{equation}
By taking the maximum value from the features of the top \(k_g\) neighboring nodes, the new feature representation for node \(i\) is derived as \(H_\text{agg}[i]\).

\subsection{Graph Scorer}
In multi-graph tasks, balancing multiple network structures is a critical challenge, as each network may contribute differently to the final representation. To address this, we propose the Graph Scorer module, which automatically learns the weight of each graph, enabling adaptive feature aggregation that reflects the varying importance of different networks.

Formally, let \(\mathbf{Z}_{g}\) denote the static graph features for each graph \(g \in \{G_{\text{dist}}, G_{\text{POI}}, G_{S}, G_{D}, G_{\text{ind}}, G_{\text{emp}}\}\). The Graph Scorer assigns each graph a softmax-normalized importance weight \(w_g\), defined as:

\begin{equation}
    w_g = \frac{\exp(s_g)}{\sum_{g' \in \mathcal{G}} \exp(s_{g'})}
\end{equation}

where \(s_g\) is a learnable scalar parameter of the graph \(g\). These weights \(w_g\) are dynamically updated during training to reflect the contribution of each graph to the overall learning process.
The weighted feature representation \(\mathbf{Z}_{\text{static}}^t\) is computed as the weighted sum of the individual graph features:
\begin{equation}
    \mathbf{Z}_{\text{static}}^t = \sum_{g \in \mathcal{G}} w_g \cdot \mathbf{Z}_{g}^t 
\end{equation}

where \(\mathbf{Z}_{\text{static}}^t\) is the aggregated representation that integrates information from all static graphs.

\subsection{Learning Objectives}
\label{sec:obj}
To ensure the learned embeddings effectively capture implicit economic relationships and dynamic changes, we design a supervised learning framework guided by complex prior knowledge. Specifically, we utilize the Barabasi Proximity, a well-established theory in complex networks, to distill meaningful information into the embeddings. The training process adopts a multi-task learning approach with two objectives: node classification to model individual city growth patterns and link prediction to incorporate inter-city industrial proximity. Joint optimization enables the model to encode both local city characteristics and inter-city economic dependencies.

\subsubsection{Node Classification of Economic Growth Rate}
The node classification task focuses on predicting the direction of economic change for each city, encompassing two key indicators: the growth rate of new companies and population employed. Specifically, it aims to determine whether the two indicators in a city from the previous year to the current year are positive or negative. We define the growth rate for both new companies and employed population first. The growth rate of new companies in city \(i\) from year \(t-1\) to year \(t\) is given by:
\begin{equation}
r_{\text{comp}} = \frac{y_{\text{comp}}^t - y_{\text{comp}}^{t-1}}{y_{\text{comp}}^{t-1}}
\end{equation}
where \( y_{\text{comp}}^t \) represents the number of new companies in city \(i\) in year \(t\), and \( y_{\text{comp}}^{t-1} \) represents the previous year.
Similarly, the growth rate of employed population \(r_{\text{emp}}\) can be calculated in the same way.

This task is crucial for capturing the temporal volatility of urban economic growth, as reflected in prior studies on dynamic economic systems~\cite{Stetsenko2021Methodical}. 
The Binary Cross-Entropy Loss (BCELoss) for this task is:
\begin{equation}
\begin{split}
\mathcal{L}_{\text{NC}} = & \alpha \cdot \text{BCELoss}(r_\text{comp,true}, r_\text{comp,pred}) \\
& +(1-\alpha) \cdot \text{BCELoss}(r_\text{emp,true}, r_\text{emp,pred})   
\end{split}
\end{equation}
where \( r_\text{true} \) and \( r_\text{pred} \) represent the ground truth and predicted labels, respectively.

\subsubsection{Link Prediction in Barabasi Network}
To incorporate complex economic relationships between cities, we define a link prediction task based on the Barabasi Proximity~\cite{barabasi2007productspace}, which captures industrial transfer patterns and economic dependencies. 
This inter-city proximity measurement involves three main stages: a) revealed comparative advantage (RCA) calculation b) proximity matrix construction c) algorithm implementation.

\paragraph{Revealed Comparative Advantage (RCA) Calculation}
For each city $c$ and industry $i$, we calculate the Revealed Comparative Advantage (RCA) as:

\begin{equation}
RCA_{c,i} = \frac{\frac{E_{c,i}}{\sum_i E_{c,i}}}{\frac{\sum_c E_{c,i}}{\sum_{c,i} E_{c,i}}}
\end{equation}

where $E_{c,i}$ represents the number of new firm registrations in industry $i$ located in city $c$. Cities with $RCA_{c,i} \geq 1$ are considered to have a comparative advantage in industry $i$.

\paragraph{Proximity Matrix Construction}
The proximity $\phi_{c,c'}$ between two cities $c$ and $c'$ is calculated using conditional probability:

\begin{equation}
\phi_{c,c'} = \min\left(P(c'|c), P(c|c')\right)
\end{equation}

where $P(c'|c)$ represents the probability that city $c'$ has RCA in an industry where city $c$ also has RCA. This symmetric measure ensures robustness to extreme values.

\paragraph{Algorithm Implementation}
The calculation process is implemented through the following key steps:

\begin{algorithm}[H]
\caption{Inter-City Proximity Calculation}
\label{alg:proximity}
\begin{algorithmic}[1]
    \State Initialize temporal range ${[}t_{\text{start}}, t_{\text{end}}{]}$
    \State Filter registrations data within specified time window
    \State Extract unique city list $\mathbf{C}$ and industry list $\mathbf{I}$
    \State Compute RCA matrix $\mathbf{R} \in \mathbb{R}^{|\mathbf{C}| \times |\mathbf{I}|}$
    \State Binarize RCA matrix $\mathbf{B}$ where $B_{c,i} = \mathbb{I}(RCA_{c,i} \geq 1)$
    \State Initialize proximity matrix $\Phi \in \mathbb{R}^{|\mathbf{C}| \times |\mathbf{C}|}$
    
    \For{each city pair $(c_i, c_j) \in \mathbf{C} \times \mathbf{C}$}
        \State Compute co-occurrence: $N_{ij} = \sum_k B_{i,k} \cdot B_{j,k}$
        \State Compute conditional probabilities:
        \State $P(c_j|c_i) = \frac{N_{ij}}{\sum_k B_{i,k}}$
        \State $P(c_i|c_j) = \frac{N_{ij}}{\sum_k B_{j,k}}$
        \State $\Phi[i,j] \leftarrow \min(P(c_j|c_i), P(c_i|c_j))$
    \EndFor
    
    \State Construct Maximum Spanning Tree (MST) from $\Phi$
    \State Remain the edges with $\Phi >= \Phi'$
\end{algorithmic}
\end{algorithm}
The Maximum Spanning Tree construction guarantees that all cities remain connected through their most significant industrial relationships, while the threshold filtering removes weak connections that may introduce noise to the link prediction task.
Through this construction process, we obtain a binary adjacency matrix where 
each entry indicates whether two cities share significant industrial proximity (presence or absence of an edge). 
%In our study, through grid search, we set $\Phi' = 0.6$ to achieve optimal balance between maintaining network sparsity and capturing meaningful economic relationships.

This task directly supervises the embeddings to reflect inter-city industrial dynamic space as a key aspect of the economic landscape. The BCELoss for this task is denoted as:
\begin{equation}
    \mathcal{L}_{\text{LP}} = \text{BCELoss}(e_\text{true}, e_\text{pred})
\end{equation}
where \( e_\text{true} \) and \( e_\text{pred} \) represent the ground truth and predicted edge weights, respectively.

\subsubsection{Combined Learning Objective}
To balance the contributions of the node classification and link prediction tasks, we introduce a weighted loss function:
\begin{equation}
  \mathcal{L} = \lambda \cdot \mathcal{L}_{\text{NC}} + (1 - \lambda) \cdot \mathcal{L}_{\text{LP}}
\end{equation}

where \( \lambda \) is a hyperparameter that controls the relative importance of each task. 
% This combined objective enables the model to distill information from both tasks, 
% ensuring effective encoding of urban economic dynamics and inter-city relationships.
% Our framework leverages prior knowledge and training objectives to learn embeddings that capture urban economic growth and dependencies.

\subsection{Model Pipeline}
The proposed model, ECO-GROW, as illustrated in Figure~\ref{fig:model}(a), is designed to predict city-level economic vitality by learning distributed embeddings that capture both static and dynamic characteristics of cities. The model takes as input a static graph of the year \(t\) and a series of dynamic graphs representing the past 15 years. 
After training, ECO-GROW generates implicit city-level representations for each year as a basis for downstream tasks.

For the static graphs, ECO-GROW utilizes six static networks. DTKGCN is applied to $G_\text{dist}^t$, $G_\text{POI}^t$, $G_\text{S}^t$, $G_\text{D}^t$, and $G_\text{ind}^t$ to capture multi-dimensional similarities by aggregating information from the most relevant neighbors. For $G_\text{emp}^t$, a GAT layer~\cite{velickovic2018graph} is used to automatically learn city relationships without predefining correlations. 
This process results in an embedding $\mathbf{Z}_g^t$ for each static graph.
To aggregate these embeddings, ECO-GROW uses a weighted combination learned through a Graph Scorer, forming the final static representation $\mathbf{Z}_\text{static}^t$.

For dynamic graphs, ECO-GROW focuses on the temporal evolution of the industry network $\{ G_\text{ind}^{t-14}, \dots, G_\text{ind}^t \}$ over the past 15 years.
Each graph in the sequence is processed by a DTKGCN layer to extract time-step features $\mathbf{Z}_{\text{dyn}}^\tau$, where $\tau \in [t-14, t]$.
These temporal features are then passed through LSTM~\cite{Gers2000Learning} to model historical dependencies and produce dynamic representations:
\begin{equation}
    \mathbf{H}_{\text{dyn}}^t = \text{LSTM}([\mathbf{Z}_{\text{dyn}}^{t-14}, \dots, \mathbf{Z}_{\text{dyn}}^t])    
\end{equation}

The static representation $\mathbf{Z}_{\text{static}}^t$ and the dynamic representation $\mathbf{H}_{\text{dyn}}^t$ are concatenated with explicit node-level features ${F}^t$ to form a unified embedding for all cities at \(t\):
\begin{equation}
    \mathbf{\mathcal{E}}^t = \text{ReLU}([\mathbf{Z}_{\text{static}}^t, \mathbf{H}_{\text{dyn}}^t, {F}^t])
\end{equation}

ECO-GROW optimizes the joint learning objective of node classification and link prediction. Once training converges, the model retains the learned embeddings $\mathcal{E}^t$ as an implicit representation of city-level economic growth.

\section{Experiments}
To evaluate the effectiveness of the learned economic growth embeddings in year $t$, we conduct two downstream regression tasks: one using the number of new companies in year \( t+1 \) as the target label, and the other using employment population in year \( t+1 \) as the target label. The embedding for each city in year \( t \) is concatenated with explicit city features (e.g., GDP, population) to form a combined representation, which is fed into two separate regression models to do predictions.
We calculate all metrics to ensure robust and generalized evaluations of our model's performance.
The experiments intend to answer the following questions: 
\begin{itemize}
    \item \textbf{RQ1:} Does our framework (ECO-GROW) improve the performance and robustness compared to the baseline models?
    \item \textbf{RQ2:} Is each component of the model effective to ECO-GROW?
    \item \textbf{RQ3:} Is ECO-GROW sensitive to the hyperparameters of different components?
\end{itemize}
%Q1. Does our framework (SAUP) improve the performance of backbones at a convincing level? Q2. Is each component of the model effective to SAUP? Q3. Is SAUP adequately robust? Q4. Is SAUP sensitive to the hyperparameters of different components? Q5. Can SAUP be trained efficiently?}

\subsection{Experiment Settings}
\subsubsection{Data Description}
We collect four real-world datasets, which focus on 297 cities across China. 
(1) \textbf{City Features}\footnotemark[2]: This city-level information includes economic indicators like GDP, population, and employment, from the year 2005 to 2022. 
(2) \textbf{Entrepreneurship data}: This dataset includes information about the registration of startups across time, cities, and industries, from the year 2005 to 2022. 
(3) \textbf{POI data}\footnotemark[3]: This dataset provides location-based information about Points of Interest in Chinese cities, from the year 2019 to 2021. 
(4) \textbf{Mobility data}\footnotemark[4]: This dataset includes information about population movement between Chinese cities, from the year 2019 to 2021.

These datasets are utilized as explicit representations, information for computing edge weights, and ground truth labels in our framework. Specifically, for the two prediction tasks, we focus on the Entrepreneurship data and the employment data from the Explicit Features dataset. To ensure consistency across datasets, we align the data for each city and year in the four datasets.

\footnotetext[2]{\url{https://www.stats.gov.cn/english}}
\footnotetext[3]{\url{https://lbsyun.baidu.com}}
\footnotetext[4]{\url{https://report.amap.com/migrate/page.do}}

% \begin{table}[h!]
%     \scriptsize
%     \caption{Data description.}
%     \begin{tabular}{@{}ccc@{}}
%     \toprule
%     \textbf{Dataset}                                                                                                                                    & \textbf{Time Range} \\ \midrule
%     Explicit Features     & {[}2005, 2021{]}    \\ \midrule
%     Entrepreneurship data                            & {[}2005, 2022{]}    \\ \midrule
%     POI data                    & {[}2019, 2021{]}      \\ \midrule
%     Mobility data                                                                                                         & {[}2019, 2021{]}      \\ \bottomrule
%     \end{tabular}
%     \label{tab:data}
% \end{table}

\subsubsection{Implementation Details}
Our proposed model is implemented using PyTorch and PyTorch Geometric. All experiments are conducted on a server equipped with Intel Xeon Gold 5218 CPU @ 2.30GHz (128 cores), 1TB RAM, and NVIDIA RTX 3090 24GB GPU. After parameter tuning, the embedding size of each city is set to \(d = 16\), and the parameters that control the loss value of the learning objectives are set to \(\alpha = 0.6\) and \(\lambda = 0.5\) respectively. We use a learning rate of 0.01 and train each baseline model for 250 epochs.
We use grid search to determine the optimal value of the L1 normalization weight in the Lasso regression model~\cite{Tibshirani1996RegressionSA}, within the range of $\{1\times 10^{-3},1\times 10^{-4},1\times 10^{-5},1\times 10^{-6}\}$. We calculate all metrics using a K-Fold cross-validation approach with \(K=5\), ensuring robust and generalized evaluations of our model's performance.

%\subsubsection{Evaluation Metrics}
%We evaluate the performance of our regression models using three metrics: Root Mean Square Error (RMSE), Mean Absolute Error (MAE), and the coefficient of determination (\textbf{$R^2$}).
%RMSE penalizes large errors, making it particularly useful for detecting significant deviations. In contrast, MAE measures the average prediction error intuitively. \(R^2\) evaluates the overall fit of the model. 
%The higher the value (the closer to 1), the stronger the predictive ability.

\subsubsection{Baseline Algorithms}
% ---------------------------------------------------------- 3 feats Result 1 ---------------------------------------------------------- %
\begin{table*}[htbp!]
    \vspace{-3mm}
    \centering
    \scriptsize
    \caption{Prediction errors and goodness of fit for forecasting the number of new companies across Chinese cities. Results for year $t$ use data up to $t$ to predict companies in $t+1$. The best result is in \textbf{bold} and the runner-up is \underline{underlined}.}
    \vspace{-2mm}
    \begin{tabular}{@{}lccccccccc@{}}
        \toprule
        \textbf{Metric} 			& \textbf{Year} & \textbf{Explicit Features}  & \textbf{Moving Window} 	& \textbf{node2vec} 	& \textbf{GAE} 	& \textbf{HREP} 	& \textbf{MVURE} 	& \textbf{MGFN} 	& \textbf{ECO-GROW}\\ 
        \midrule
        \multirow{4}{*}{RMSE↓}      & 2019 			& \underline{16726.499}  		     & 23619.512  		        & 21843.818  			& 22107.894  	& 19328.523  		& 18787.233  		& 17165.253    		& \textbf{16029.035  } \\ 
        						& 2020 			& 21012.320  		     & 30559.153  		        & 22812.730  			& 22510.601  	& 22636.571  		& \underline{20301.098}  		& 20978.207   		& \textbf{15303.907  } \\ 
       						  & 2021 		  & 23924.155  		       & 28231.584  		      & 24909.030  		      & 30433.218  	  & \underline{23576.446}  	      & 26243.253  		  & 23869.707    	  & \textbf{17710.724  } \\ 
        						& Average 		& \underline{20554.324}  		     & 27470.083  		        & 23188.526  			& 25017.238  	& 21847.180  		& 21777.195  		& 20671.056   		& \textbf{16347.888  } \\ 
        \midrule
        \multirow{4}{*}{MAE↓} 	    & 2019			& 8500.289      & 11628.066  		        & 11954.124  			& 10766.108  	& 11067.699  		& 10592.544  		& \underline{8370.484}  		& \textbf{8199.070  }   \\ 
        						& 2020 			& \underline{11490.189}  			 & 15397.325  		        & 12919.301  			& 13920.786  	& 13889.582  		& 11678.060  		& 12029.129  		& \textbf{9439.828  } \\ 
       						  & 2021 		  & 16923.181  		       & 19163.416  		      & 17698.153  		      & 18496.387  	  & \underline{16355.492}  	      & 18038.560    		  & 16933.148  		  & \textbf{12332.495  } \\ 
        						& Average 		& \underline{12304.553}  		     & 15396.269  		        & 14190.526  			& 14394.427  	& 13770.924  		& 13436.401  		& 12444.254   		& \textbf{9990.464  } \\ 
        \midrule		
        \multirow{4}{*}{R\(^2\)↑}   & 2019 		    & \underline{0.851}  			     & 0.682  			        & 0.441  				& 0.776  		& 0.838 			& 0.817  			& 0.824  			& \textbf{0.855  } \\ 
        						& 2020 		    & 0.849  			     & 0.670  			        & 0.811  				& 0.828  		& 0.817  			& \underline{0.861}  			& 0.844  			& \textbf{0.909  } \\ 
       							& 2021 		    & 0.817  			     & 0.763  			        & 0.801  				& 0.728  		& \underline{0.826}  			& 0.785  			& 0.819  			& \textbf{0.905  } \\ 		
        						& Average 	    & \underline{0.839}  			     & 0.705  			        & 0.684  				& 0.778  		& 0.827  			& 0.821  			& 0.829       		& \textbf{0.889  } \\ 
        \bottomrule
    \end{tabular}
    \label{tab:results_company}
    \vspace{-3mm}
\end{table*}

\begin{table*}[htbp!]
    \centering
    \scriptsize
    \caption{Prediction errors and goodness of fit for forecasting population employed across Chinese cities. Results for year $t$ use data up to $t$ to predict companies in $t+1$.}
    \vspace{-2mm}
    \begin{tabular}{@{}lccccccccc@{}}
        \toprule
        \textbf{Metric} 			& \textbf{Year} & \textbf{Explicit Features}  & \textbf{Moving Window} 	& \textbf{node2vec} 	& \textbf{GAE} 	& \textbf{HREP} 	& \textbf{MVURE} 	& \textbf{MGFN} 	& \textbf{ECO-GROW}\\ 
        \midrule
        \multirow{4}{*}{RMSE↓}      & 2019 			& 12.375  		     & 17.310  		        & 12.217  			& 11.920  	& 12.476  		& 11.615  		& \underline{11.030}    		& \textbf{11.025  } \\ 
        						& 2020 			& 16.850  		     & 20.274  		        & 14.851  			& 14.933  	& 17.073  		& 15.882  		& \underline{14.414}   		& \textbf{13.968  } \\ 
       						  & 2021 		  & \underline{15.016}  		       & 17.560  		      & 17.245  		      & 16.205  	  & 16.514  	      & 15.360  		  & 16.039    	  & \textbf{14.196  } \\ 
        						& Average 		& 14.747  		     & 18.381  		        & 14.771  			& 14.353  	& 15.355  		& 14.286  		& \underline{13.827}   		& \textbf{13.063  } \\ 
        \midrule
        \multirow{4}{*}{MAE↓} 	    & 2019			& 6.418      & 6.950  		        & 6.053  			& 5.679  	& 7.327  		& 4.913  		& \underline{4.652}  		& \textbf{4.589  }   \\ 
        						& 2020 			& 6.552  			 & 7.274  		        & 6.134  			& 6.064  	& 8.577  		& 6.534  		& \underline{5.447}  		& \textbf{5.368  } \\ 
       						  & 2021 		  & 8.463  		       & \underline{7.445}  		      & 8.155  		      & 8.457  	  & 8.516  	      & 7.540  		  & 7.705  		  & \textbf{6.858  } \\ 
        						& Average 		& 7.144  		     & 7.223  		        & 6.781  			& 6.733  	& 8.140  		& 6.329  		& \underline{5.935}   		& \textbf{5.605  } \\ 
        \midrule		
        \multirow{4}{*}{R\(^2\)↑}   & 2019 		    & 0.950  			     & 0.944  			        & 0.941  				& 0.954  		& 0.955  			& 0.962  			& \underline{0.964}  			& \textbf{0.965  } \\ 
        						& 2020 		    & 0.921  			     & 0.901  			        & 0.925  				& 0.927  		& 0.910  			& 0.927  			& \underline{0.935}  			& \textbf{0.937  } \\ 
       							& 2021 		    & 0.935  			     & 0.933  			        & 0.888  				& 0.927  		&  0.922  			& \underline{0.936}  			& 0.931  			& \textbf{0.944  } \\ 		
        						& Average 	    & 0.935  			     & 0.928  			        & 0.918  				& 0.936  		& 0.929  			& 0.942  			& \underline{0.944}       		& \textbf{0.949  } \\ 
        \bottomrule
    \end{tabular}
    \label{tab:results_employment}
    \vspace{-5mm}
\end{table*}

We conduct experiments on the following baseline models, categorized into two types according to whether they rely on explicit or implicit representations. All baseline models are designed with the objective of generating city embeddings, enabling a fair comparison with our model.

\noindent (1) Explicit baselines: These methods directly use observable city attributes. Yet they do not capture hidden, complex relationships across multiple city attributes.
\begin{itemize}
    \item \textbf{Explicit Features}: Uses raw indicators (e.g., GDP, population, new companies, employment in year t) as input to a Lasso regression.

    \item \textbf{Moving Window}: Captures temporal dynamics by using new companies and population employed over the past 15 time steps as sequential features.
\end{itemize}

\noindent (2) Implicit baselines: These methods aim to learn latent city embeddings using graph-based or region embedding techniques. Yet they fail to capture the dynamic nature of urban systems and the broader inter-city economic relationships.
\begin{itemize}
    \item \textbf{Node2Vec}~\cite{grover2016node2vec}: 
    Learns node embeddings by performing biased random walks on the graph.
    
    \item \textbf{Graph Autoencoder (GAE)}~\cite{kipf2016variational}: 
Encodes city graphs into latent embeddings by reconstructing graph topology.
    
    \item \textbf{MVURE}~\cite{zhang2021multi}, \textbf{MGFN}~\cite{wu2022multi}, \textbf{HREP}~\cite{zhou2023heterogeneous}: 
    Multi-graph or heterogeneous region embedding frameworks that fuse multiple urban networks through attention or adaptive weighting. HREP represents the current state-of-the-art in multi-graph region embedding.
\end{itemize}

For explicit baselines, we use features directly as input for downstream tasks. For implicit baselines, we apply recommended hyperparameters, input static networks from year $t$, and build task-specific objectives. Each model is trained for 250 epochs to produce a 16-dimensional embedding, which is then combined with explicit features for evaluation.

\subsection{Result Analysis (\textbf{RQ1})}

%%%%%%%%%%%%%%%%%%% Figure: model %%%%%%%%%%%%%%%%%%%
% \begin{figure}[h!]
% % \centering
% \includegraphics[width=0.5\textwidth]{figure/stat/wuhan_vs_national_growth.pdf}
% \caption{This figure displays the number of new companies (blue bars) and the growth rate (red lines) in Wuhan (a) and all cities in China (b) from 2019 to 2022.}
% \label{fig:shock}
% \end{figure}
%%%%%%%%%%%%%%%%%%% Figure: model %%%%%%%%%%%%%%%%%%%
Table~\ref{tab:results_company} and~\ref{tab:results_employment} list the performance of all models in 2019, 2020, and 2021, as well as their average results for predicting the number of new companies and the employment population respectively.
It is important to note that the result for $t$ means that we use the data from before and $t$ to predict the number of new companies in year $t+1$.

\subsubsection{Performance Across Tasks}
For both prediction tasks, new companies and employment population, ECO-GROW consistently outperforms all baselines in RMSE, MAE, and \(R^2\).
Specifically, ECO-GROW reduces RMSE and MAE by 20.46\% and 18.81\% respectively, and improves \(R^2\) by 5.96\% in the first task. 

Although Explicit Features achieves the second-best performance for entrepreneurship and MGFN ranks second for employment, their improvements are limited due to their reliance on static attributes or single-view supervision. The moving window method has inherent limitations in capturing long-term trends and dynamic changes in economic systems. Node2Vec and GAE, on the other hand, rely heavily on fixed neighborhood structures and static graph representations.
In contrast, ECO-GROW utilizes complex prior knowledge alongside dynamic information to better model temporal trends in economic growth. 
% As a result, ECO-GROW demonstrates superior performance in most evaluation scenarios, underscoring its effectiveness in modeling economic growth under diverse conditions.

\subsubsection{Robustness Across Tasks}
While some baselines perform well on individual tasks, they lack consistency across both. ECO-GROW maintains stable superiority in all evaluations, demonstrating strong generalization.
To better understand the complexity and stability of the two downstream tasks, we calculate the KL divergence between yearly distributions: smaller values indicate greater stability and simplicity. 

\begin{figure}[htbp]
% \vspace{-3mm}
\centering
\includegraphics[width=0.7\linewidth]{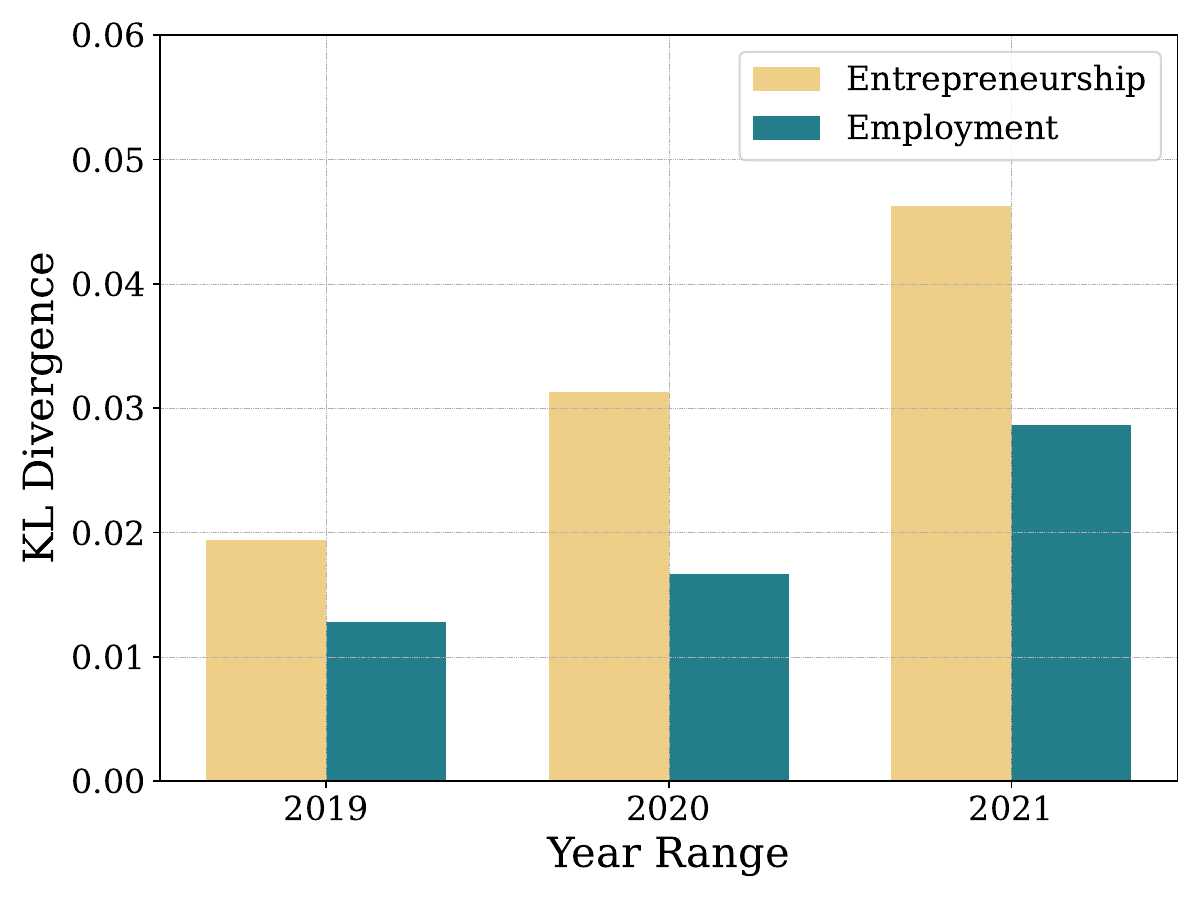}
\vspace{-3mm}
\caption{KL Divergence between years for two downstream tasks. For year $t$, the result represents the distribution shift for each task from the previous year $t-1$ to the current year $t$.}
\label{fig:kl}
\vspace{-3mm}
\end{figure}

As shown in Figure~\ref{fig:kl}, entrepreneurship prediction exhibits a higher KL divergence compared to the employment prediction in all years, suggesting that the former task is more challenging than the latter. ECO-GROW shows solid improvements in both tasks, with a more substantial enhancement in the challenging one, demonstrating its superior capability to handle more complex urban economic indicators.
%%%%%%%%%%%%%%%%%%% Figure: kl %%%%%%%%%%%%%%%%%%%
%%%%%%%%%%%%%%%%%%% Figure: kl %%%%%%%%%%%%%%%%%%%

\subsubsection{Robustness Across Years}
Although HREP and MVURE perform well in the year 2021 and year 2020 respectively, ECO-GROW consistently outperforms all baseline models in both tasks across all three years (2019, 2020, and 2021), demonstrating its ability to maintain strong performance over time. This stable performance further underscores the robustness of ECO-GROW, showing that it is not sensitive to annual variations in the data and can be relied upon for forecasting across different periods.

\subsection{Ablation Analysis (\textbf{RQ2})}
We conduct three ablation experiments focusing on both downstream tasks and the results are shown in Tabel~~\ref{tab:abl_comp} and Table~\ref{tab:abl_emp} respectively. For the entrepreneurship prediction task,  replacing the DTKGCN layer with a standard GCN (w/o DTKGCN) increases RMSE by 18.14\% on average. Removing the Graph Scorer (w/o GS) and directly concatenating graph features raises \(R^2\) by 6.52\%, underscoring the importance of learning graph weights. Eliminating the dynamic graph input and LSTM module (w/o LSTM) leads to a 9.99\% MAE increase, demonstrating their necessity for capturing temporal dynamics. Similarly, for employment trends prediction, the ablation experiments demonstrate a similar pattern of results.
The full model consistently surpasses ablated versions across all metrics, confirming the effectiveness of its key components.
\begin{table}[htbp!]
    \vspace{-2mm}
    \scriptsize
    % \tiny
    \caption{Ablation results for predicting the number of new companies.}
    \vspace{-2mm}
    \begin{tabular}{@{}lccccc@{}}
        \toprule
        \textbf{Metric} 		& \textbf{Year} 	& \textbf{w/o DTKGCN} 	& \textbf{w/o GS} 	& \textbf{w/o LSTM} 	& \textbf{ECO-GROW}\\ 
        \midrule
        \multirow{4}{*}{RMSE↓}  & 2019 				& 19816.097   			& 26966.652    				& 17269.829    			& \textbf{16029.035 } \\ 
        				    & 2020 				& 18804.240           	& 16771.433    				& 20732.783    			& \textbf{15303.907 } \\ 
       						& 2021 				& 19317.750    			& 20401.466    				& 18571.407    			& \textbf{17710.724 } \\ 
        					& Average 			& 19312.696    			& 21379.850    				& 18858.007    			& \textbf{16347.888 } \\ 
        \midrule
        \multirow{4}{*}{MAE↓} 	& 2019 				& 9315.993    			& 11335.491    				& 8781.204    			& \textbf{8199.070 } \\ 
        					& 2020 				& 11244.288            	& 10570.642    				& 11465.346   			& \textbf{9439.828 } \\ 
       						& 2021 				& 13630.000    			& 13118.438    				& 12718.232    			& \textbf{12332.495 } \\ 
        					& Average 			& 11396.760    			& 11674.857    				& 10988.261    			& \textbf{9990.464 } \\ 
        \midrule
        \multirow{4}{*}{R\(^2\)↑}  & 2019 			& 0.802    				& 0.719    					& 0.830    				& \textbf{0.855 } \\ 
        					    & 2020 			& 0.854          		& 0.896    					& 0.861    				& \textbf{0.909 } \\ 
       					        & 2021 			& 0.889   				& 0.878    					& 0.896    				& \textbf{0.905 } \\ 
        						& Average 		& 0.848    				& 0.831   				    & 0.862    	      		& \textbf{0.889 } \\ 
        \bottomrule
    \end{tabular}
    \label{tab:abl_comp}
    \vspace{-5mm}
\end{table}

\begin{table}[htbp!]
    \scriptsize
    % \tiny
    \caption{Ablation results for predicting population employed.}
    \vspace{-2mm}
    \begin{tabular}{@{}lccccc@{}}
        \toprule
        \textbf{Metric} 		& \textbf{Year} 	& \textbf{w/o DTKGCN} 	& \textbf{w/o GS} 	& \textbf{w/o LSTM} 	& \textbf{ECO-GROW}\\ 
        \midrule
        \multirow{4}{*}{RMSE↓}  & 2019 				& 11.474   			& 11.977    				& 12.092    			& \textbf{11.025 } \\ 
        				    & 2020 				& 14.809          	& 31.886    				& 14.651    			& \textbf{13.968 } \\ 
       						& 2021 				& 14.607    			& 17.866    				& 14.410    			& \textbf{14.196 } \\ 
        					& Average 			& 13.630    			& 20.576    				& 13.718    			& \textbf{13.063 } \\ 
        \midrule
        \multirow{4}{*}{MAE↓} 	& 2019 				& 5.440    			& 5.436    				& 5.485    			& \textbf{4.589 } \\ 
        					& 2020 				& 6.557            	& 9.970    				& 6.212   			& \textbf{5.368 } \\ 
       						& 2021 				& 7.246    			& 8.073    				& 6.920    			& \textbf{6.858 } \\ 
        					& Average 			& 6.414    			& 7.826    				& 6.206    			& \textbf{5.605 } \\ 
        \midrule
        \multirow{4}{*}{R\(^2\)↑}  & 2019 			& 0.959    				& 0.960    					& 0.959    				& \textbf{0.965 } \\ 
        					    & 2020 			& 0.932          		& 0.840    					& 0.932    				& \textbf{0.937 } \\ 
       					        & 2021 			& 0.940   				& 0.928    					& 0.939    				& \textbf{0.944 } \\ 
        						& Average 		& 0.944    				& 0.910   				    & 0.943    	      		& \textbf{0.949 } \\ 
        \bottomrule
    \end{tabular}
    \label{tab:abl_emp}
    \vspace{-5mm}
\end{table}

\subsection{Parameters Sensitivity (\textbf{RQ3})}
\begin{figure}[htbp]
\vspace{-3mm}
\centering
\includegraphics[width=\linewidth]{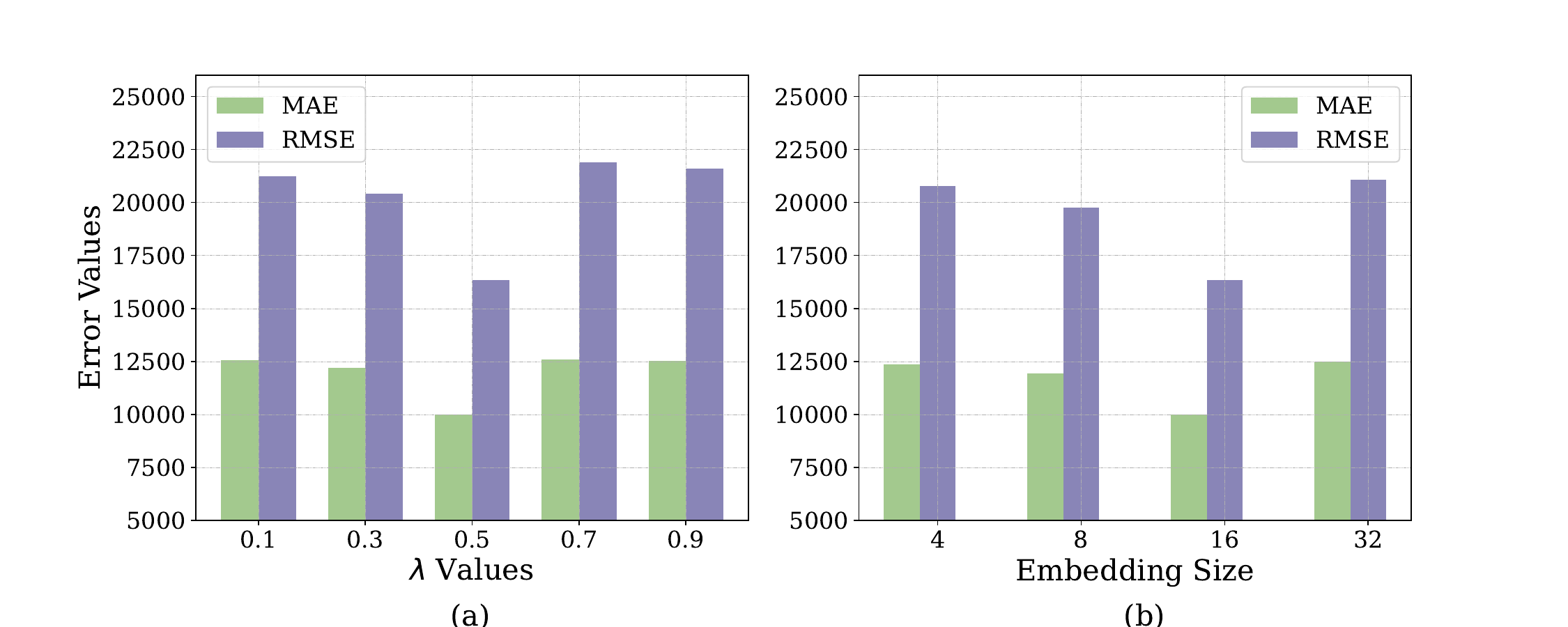}
\vspace{-6mm}
\caption{Sensitivity Analysis of the parameters $\lambda$ and $d$ in predicting the number of new companies respectively.}
\label{fig:sensitivity_all}
\vspace{-6mm}
\end{figure}

We evaluate the sensitivity of the model to two critical hyperparameters:$\lambda$, which balances the node classification and link prediction tasks, and the embedding size $d$. In the sensitivity analysis for $\lambda$, we fix $d=16$. In Figure~\ref{fig:sensitivity_all} (a), we observe that, as $\lambda$ increases from 0.1 to 0.5, there is a decrease in RMSE and MAE. However, when $\lambda$ exceeds 0.5, performance begins to deteriorate. This pattern shows that the model leverages both the node classification task (which focuses on individual city growth patterns) and the link prediction task (which models the inter-city dependencies) effectively.

Similarly, for the sensitivity analysis for embedding size, we fix $\lambda = 0.5$. The results, shown in Figure~\ref{fig:sensitivity_all} (b), indicate that an embedding size of $d = 16$ achieves the best performance, balancing the model's ability to represent features with its complexity, while larger embedding sizes lead to overfitting and reduced generalization.

% %%%%%%%%%%%%%%%%%%% Figure: sensitivity_lambda %%%%%%%%%%%%%%%%%%%
% \begin{figure}[htbp]
% \centering
% \includegraphics[width=\linewidth]{figure/sensitivity/sensitivity_lambda.pdf}
% \caption{Sensitivity Analysis of the $\lambda$ parameter in predicting the number of new companies.}
% \label{fig:sensitivity_lambda}
% \end{figure}
% %%%%%%%%%%%%%%%%%%% Figure: sensitivity_lambda %%%%%%%%%%%%%%%%%%%
% %%%%%%%%%%%%%%%%%%% Figure: sensitivity_embedding_size %%%%%%%%%%%%%%%%%%%
% \begin{figure}[htbp]
% \centering
% \includegraphics[width=\linewidth]{figure/sensitivity/sensitivity_embedding_size.pdf}
% \caption{Sensitivity Analysis of the embedding size $d$ in predicting the number of new companies.}
% \label{fig:sensitivity_embedding_size}
% \end{figure}
% %%%%%%%%%%%%%%%%%%% Figure: sensitivity_embedding_size %%%%%%%%%%%%%%%%%%%
%%%%%%%%%%%%%%%%%%% Figure: sensitivity_all %%%%%%%%%%%%%%%%%%%
%%%%%%%%%%%%%%%%%%% Figure: sensitivity_all %%%%%%%%%%%%%%%%%%%
% \begin{figure}[htbp]
% \centering
% \begin{minipage}[b]{0.45\linewidth}
%     \centering
%     \includegraphics[width=\linewidth]{figure/sensitivity/lambda.pdf}
%     \caption{Sensitivity Analysis of the $\lambda$ parameter in predicting the number of new companies.}
%     \label{fig:sensitivity_lambda}
% \end{minipage}
% \hfill
% \begin{minipage}[b]{0.45\linewidth}
%     \centering
%     \includegraphics[width=\linewidth]{figure/sensitivity/embedding_size.pdf}
%     \caption{Sensitivity Analysis of the embedding size $d$ in predicting the number of new companies.}
%     \label{fig:sensitivity_embedding_size}
% \end{minipage}
% \end{figure}

\section{Conclusion}
This paper introduces ECO-GROW, a multi-graph framework designed to model and predict urban economic vitality across Chinese cities. By integrating industrial, regional, and mobility similarities, as well as temporal dynamics, ECO-GROW effectively captures the complex relationships driving urban economic development. 

A key innovation of ECO-GROW is its use of complex prior knowledge to guide graph representation learning. Instead of directly predicting a value, our framework optimizes economic representations aligned with growth-related tasks, ensuring knowledge-rich and interpretable embeddings. The Graph Scorer adaptively learns the importance of static networks, while the DTKGCN dynamically selects the most relevant neighboring cities based on the characteristics of each graph type. These components enable ECO-GROW to generate accurate and robust representations.

This work demonstrates the model’s capacity to enhance urban planning and policymaking by forecasting the key economic indicators, offering a tool to guide future urban growth strategies. Beyond predicting the number of new companies and employment population, the learned representations offer potential for broader applications, such as forecasting housing market trends or analyzing urban economic patterns. By open-sourcing our code, we aim to promote further research and practical applications in urban planning and sustainable development.

\section*{Acknowledgement}
This project was supported by the Science and Technology Commission of Shanghai Municipality (No. 25692108200).

\bibliography{references}  % 指定 .bib 文件名称，这里是 references.bib

%\appendix
%\input{appendix.tex} 

\end{document}